# Encoder-Decoder Based Long Short-Term Memory (LSTM) Model for Video Captioning

Sikiru Adewale[1], Tosin Ige[2], Bolanle Hafiz Matti[3]
Department of Computer Science, Virginia Tech. SW Blacksburg, Virginia. USA[1]
Department of Computer Science, University of Texas at El Paso, Texas, USA[2]
Department of Mathematics and Statistics, Austin Peay State University, Tennessee. USA[3]

*Abstract*—This work demonstrates the implementation and use of an encoder-decoder model to perform a many-to-many mapping of video data to text captions. The many-to-many mapping occurs via an input temporal sequence of video frames to an output sequence of words to form a caption sentence. Data preprocessing, model construction, and model training are discussed. Caption correctness is evaluated using 2-gram BLEU scores across the different splits of the dataset. Specific examples of output captions were shown to demonstrate model generality over the video temporal dimension. Predicted captions were shown to generalize over video action, even in instances where the video scene changed dramatically. Model architecture changes are discussed to improve sentence grammar and correctness.

## I. Introduction

Video captioning is an important extension of image captioning. As data generation and storage capacity increases, multimodal data, such as video, increases in abundance. A necessary tool for accessibility, archival, and display is the ability to generate human-readable captions from these new data sources. Specifically, generating captions for video data that include description of action would be highly useful for automatic video labeling. Captioning video data is a more difficult task than image captioning, as video data is the concatenation of image frames across a temporal dimension. In the temporal domain, there can be many changes across the image frames, such as changes in brightness, camera angle, camera position, field-of-view, subject actions, and even the scene itself.

Video action recognition is the subfield of deep learning and computer vision that broadly describes the recognition of actions in video data. A subset of this problem is constraining video action to human actions, although the superset of all video actions certainly includes any action capable of being recorded on video. There are many different approaches to video action recognition, including 3D convolutional neural networks (CNNs) with and without attention layers, as well as multiple stream networks utilizing different video temporal segments, temporal resolutions, and even optical flow [1]. The most difficult part of video action recognition, or as described in this work as video captioning, is the temporal change in the image frames.

A basic approach to handling the temporal dimension of the image data is to treat the video stream as simply a list of temporal image frames. A one-to-many model for text generation can be highly effective at captioning individual frames. In this work, we extend image captioning to video captioning using a many-to-many architecture based on an encoder-decoder model [2]. An encoder (long short-term memory layer) LSTM is added to handle the temporal domain of the input video, while the decoder LSTM handles the text sequence output. Feature vectors are generated using a pretrained 2D CNN model to convert each frame to a feature vector. In the video domain, this feature vector is now a matrix with a temporal dimension. The goal of this project is to produce a single caption for each video that is more general than a caption for any single frame of the video. With input video data that undergoes significant scene or subject changes across the temporal dimension, the caption can be a summary or amalgamation of all frames.

## II. Related Works

Video captioning: This uses the concept of giving a video and a language query to track the moment in the video. The goal is to localize objects that are mentioned in the sentence in the video. Many works have been done in the attempt to localize a moment in the video. Video localizing aims to track a moment from an untrimmed video for a given text query. Some of the challenges in this area are the need for dense-fine regional annotations in the videos and a possible solution can be achieved by exploring weakly supervised video captioning, zero-shot video captioning, and fully supervised video captioning. How to characterize the relationships between videos and sentences is another notable challenge. Some of the previous solutions are matching spatial regions in specific frames with nouns/pronouns in the sentence but the limitation in this approach is that the spatio-temporal dynamics of the videos are not exploited. The nouns/pronouns are less expressive than a natural sentence [3]–[6].

Fully Supervised Video captioning: In fully supervised visual localization, there is need for annotation of the action in the video. No focus on precision in video moment localization and sliding window approach has fixed window and anchor based compromises on precision of moments. The boundary aware temporary language captioning and self-attention mechanism on interaction between video and language for capturing contextual information for better semantic understanding are proposed [7], [8]. Increasing temporal receptive fields, and using regression for action boundary prediction and distance of the given frame from the start and end frame [9], [10]. The fully supervised visual localization is very extensive and time-consuming to annotate the videos. In an attempt to reduce the cost of annotation, weakly supervised, unsupervised, and zero shot approaches are explored. The common evaluation metric depends on the mean value which is not robust to anomalies, extreme values, and outliers.

Weakly Supervised Video captioning: Weakly supervised method uses some unlabeled and labeled data to train a model. Fully supervised requires heavy annotations, but temporal localization (frame-wise attention), spatial interaction, hierarchical multi-instance levels for optimization are achieved through weakly supervised [11]. Co-attention was not previously used in video moment retrieval tasks until multi-level co-attention mechanics helped to improve alignment between video and text including positional encoding on frame features [12]. Decomposing the spatial and temporal representations to collect all-sided cues for precise captioning, there are interaction level counterfactual transformations of feature, temporal video captioning using video corpus for moment retrieval, and cross modal retrieval with the aid of contrastive learning [13]– [15]. Weakly supervised was able to take care of annotation due to video captioning, but the additional cost of natural language was not reduced in this approach. Using densely big networks during pretraining and fine-tuning, the size of the features is inversely proportional to the benefit of the unlabeled data derived from the dense networks. Accuracy was used in evaluating the performance and the metric is not appropriate for sparse data [16].

Unsupervised Video captioning: Unsupervised visual localization uses the unsupervised model to establish a location of a visual object with respect to the corresponding text based data. A method was proposed to map languages through the visual domain using only unpaired instructional videos in paper [5]. Prototypical contrastive learning is an unsupervised method that connects clustering and contrastive method to encrypt semantic structures. This is obtained by clustering into the grounded space after learning low-level labels for the aim of object contrasting [6]. Some of the problems like clustering similar activities and manually-captioned videos lacking some activities that are not captioned are addressed by the prototypical contrastive learning model which maps languages with the unpaired videos as input. The problem of connecting words in different languages by visual object is tackled with unsupervised method.

The strengths of unsupervised visual localization include fewer data captioning for training the model, reduction in annotation cost, and time efficiency in the training. Self supervised learning is a method that uses unlabeled data to build a model [3], [4] and it can be a sub-division of unsupervised method. Most of the models used accuracy as evaluation metrics which does not perform well in a situation of imbalanced data.

III. METHODS

The data used for this project was the Microsoft Research Video Description Corpus (MSVD) [17]. The MSVD dataset contains 1970 video files from YouTube with their text descriptions. Each video file had multiple text descriptions, typically 20 to 40 labeled sentences describing each video. The dataset was split into 3 sets: training, validation, and test. The training set was taken to be 90% of the dataset, with 5% held out for validation and 5% held out for test. Thus, the training dataset contained 1773 videos, while the validation and test datasets contained 99 and 98 videos, respectively.

The labeled text descriptions were stored in a text file containing the video file name and the nested list of its descriptions. Thus, this file was parsed, and a description dictionary was created to map keys (video filenames) to their descriptions (nested list of sentence captions). During parsing, each sentence description was prepended and appended with unique tokens to denote the beginning and end of the description sentence. Also during parsing, sentence descriptions with length less than six words or greater than ten words were not added to the description dictionary. Figure 1 shows the histogram of word lengths in the final description dictionary. As descriptions outside of the desired word length range were not considered, only descriptions between six and ten words exist in the dictionary. Most descriptions in the dictionary had eight words.

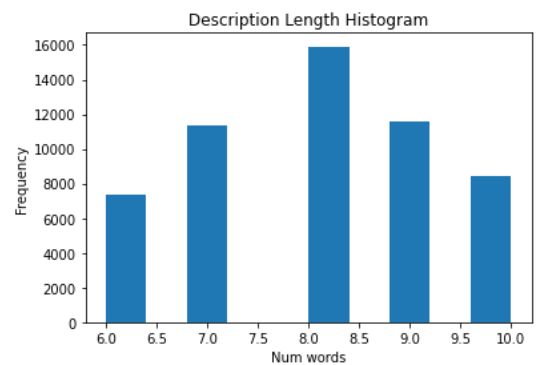

Fig. 1: Description length histogram.

After associating each video in the entire dataset with its filtered-length descriptions, the list of video keys (filenames) were randomly shuffled and split into the training, validation, and test lists. To create the vocabulary of the corpus, a Keras tokenizer was created to parse the descriptions in the *training* set only. Limiting the vocabulary generation to only the training set descriptions ensures the model only knows words associated with the training videos. The tokenizer was used to generate a vocabulary of 1500 words, so only the most frequently occurring 1500 words in the training set were kept. Figure 2 shows the thirty most commonly occurring words (tokens) in the training set from the tokenizer. The most commonly occurring token is "a", while the next two most commonly occurring words are "bos" and "eos", the beginning and end of string tokens.

The tokenizer also creates dictionaries to map the words in the vocabulary to numerical indices ("a" goes to 1, "bos" goes to 2, etc.) and also indices back to words (1 back to "a"). Figure 3 shows the tokenizer output on a training sentence, where the sentence (top line) is mapped to a list of token indices (bottom line).

decimation ensures that each video had the same dimension in the time domain. Each frame was then passed to the VGG16 CNN model to extract a 4096 long feature vector for each frame. Thus, each video was associated with a feature matrix of size 80x4096, and this data was stored in a dictionary associated with the video keys (filenames). Figure 4 shows an example feature matrix for a video in the training set. The matrix rows are the image spatial feature vectors extracted from the VGG16 model, while the image columns represent the temporal dimension of the video. An abrupt change from row to row in the feature matrix corresponds to a scene change in the video.

With the data preprocessed, the model was then constructed. The encoder-decoder model for video captioning using two LSTMs to perform sequence-to-sequence (or many-to-many) mapping between the input and output data. The encoder model encodes video temporal information using the video feature matrices, while the decoder part of the model takes the one-hot-encoded text data to generate output words. Figure 5

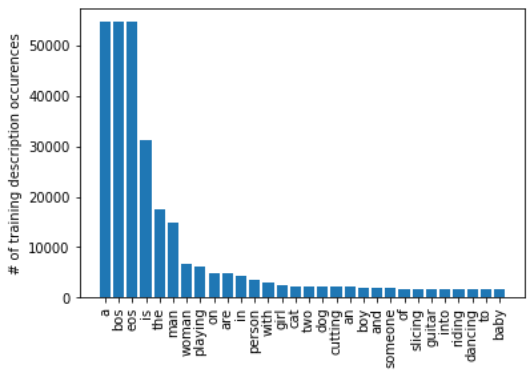

Fig. 2: Vocabulary histogram.

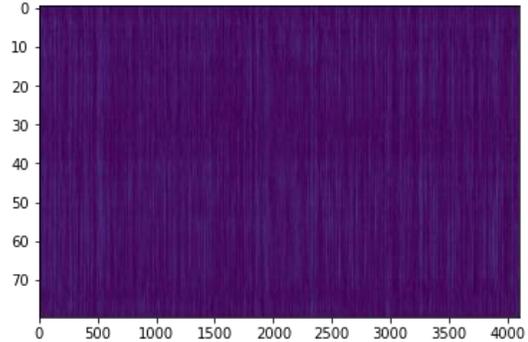

Fig. 4: Example video feature matrix.

Fig. 3: Sentence tokenization

For data generation for the encoder-decoder training model, the sentence descriptions were padded to the same maximum length of ten words, and the sequence indices were converted to one-hot encoded (categorical) vectors. Thus, each sentence description was a matrix of size 10x1500. These descriptions form the output of the decoder training model. The encoder training model takes the video spatial and temporal information. The VGG16 pre-trained model was used from Keras to extract feature vectors from each video frame. For each video, all of its temporal frames were extracted using OpenCV, resized spatially, and stored in a temporal list. As each video had a different number of frames, a linearly spaced vector of 80 frame indices were generated and mapped to the length of each video. For example, a video of length 123 frames would have 80 frames extracted, where the zeroth frame was the zeroth frame and the 80th frame was the 123th frame. This temporal shows the constructed model, where the encoder and decoder LSTMs each have a latent dimension of 512 and their output shape is the temporal dimension of each dataset. In other words, the 80 temporal steps of the encoder (80 video frames) and the 10 temporal steps of the decoder (10 word sentences) form the output shape along with the latent dimensions. Only the last temporal LSTM cell is taken from the encoder model, as its hidden and cell states contain information of all 80 temporal inputs. The decoder LSTM is then connected to a dense layer with softmax activation function to output the one hot-encoded target vector for the next predicted word in the caption.

```
Model: "model"
_________________________________________________________________________________
Layer (type)          Output Shape              Param #    Connected to
=================================================================================
encoder_inputs (InputLayer)  [(None, 80, 4096)]  0          []

decoder_inputs (InputLayer)  [(None, 10, 1500)]  0          []

endcoder_lstm (LSTM)   [(None, 80, 512),         9439232    ['encoder_inputs[0][0]']
                        (None, 512),
                        (None, 512)]

decoder_lstm (LSTM)    [(None, 10, 512),         4122624    ['decoder_inputs[0][0]',
                        (None, 512),                         'endcoder_lstm[0][1]',
                        (None, 512)]                         'endcoder_lstm[0][2]']

decoder_relu (Dense)   (None, 10, 1500)          769500     ['decoder_lstm[0][0]']
=================================================================================
Total params: 14,331,356
Trainable params: 14,331,356
Non-trainable params: 0
_________________________________________________________________________________
```

Fig. 5: Training model.

The model was trained using a data generator function which, for each training epoch, loops through all video keys in the training data set, extracts the feature matrix from the feature dictionary, and extracts the padded and one-hot encoded description matrices from the description dictionary. For each token in the description matrix (text decoder input), the input and target for the decoder are extracted by taking the last token as the output target and all previous tokens as the decoder inputs. These iterations are counted, and when the iterations equal the batch size, yielded to the model for fitting. A batch size of 50 was used to train the model for 80 epochs using the Adam optimizer with a learning rate of 0.0001 And categorical cross-entropy loss. Loss and accuracy metrics were recorded for the training and validation datasets during training. Figure 6 shows the categorical cross-entropy loss of the model during training, while Figure 7 shows the categorical accuracy.

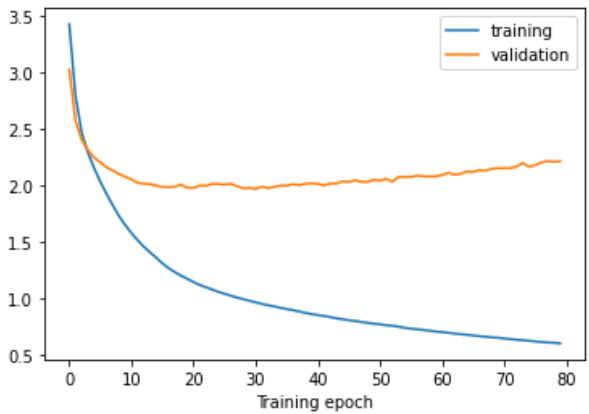

Fig. 6: Model training loss.

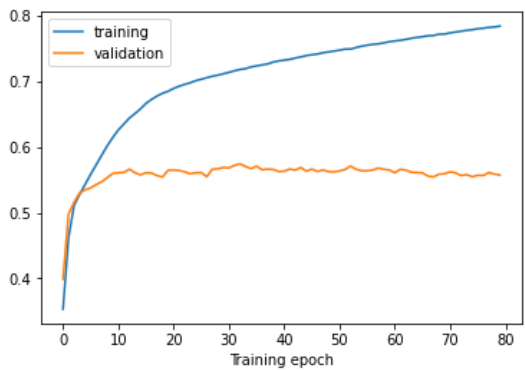

Fig. 7: Model training accuracy.

With the encoder-decoder model architecture, the training and inference models are different. The encoder model is the same for training and inference, as only the state of the last cell of the encoder LSTM is used in both cases. During training, the encoder and decoder models see both inputs (temporal video and text data) simultaneously to update the model weights with respect to both inputs. However, during inference, the text decoder model only uses as input the final temporal state of the video encoder model. Thus, during inference, the encoder and decoder models are used separately. The encoder model takes the 80x4096 video feature matrix for each frame and outputs only the hidden and cell state of the last LSTM layer. These states are fed to the text decoder model simultaneously with the text token. To predict captions, the first text token is the "bos" token to denote the start of the string. The index of the maximum of the one-hot-encoded output vector (greedy search) is taken and converted to a text word using the tokenizer. The inference model is applied in a loop to generate successive tokens from the "bos", appended each output token to the input sentence, and using the previous output as the next iteration's input. The loop stops when the "eos" token is predicted, denoting the end of the sentence, or the maximum length of the description (set as ten words) is reached.

IV. EXPERIMENTAL RESULTS

The inference model was applied to all videos in each dataset (training, validation, and test). Predicted captions were stored in a dictionary accessible via the video keys. A 2-gram BLEU score was generated to compare the predicted caption to all the labeled descriptions for each video. A scatter plot of the BLEU scores for each video in each split of the dataset is shown, as well as a histogram.

*A. Training set*

For the training set, the categorical accuracy of the model at the end of training was close to 80%, so the expectation is that the BLEU scores between predicted and labeled captions in the training set is quite good. Figure 8 shows the BLEU scores across all videos in the training set. Figure 9 shows the frequency distribution (histogram) of BLEU scores in the training set. The average BLEU score for predicted captions on the training set is 91.8%, which is quite high than the state of-art [18] which is 57.8% for BLEU-4.

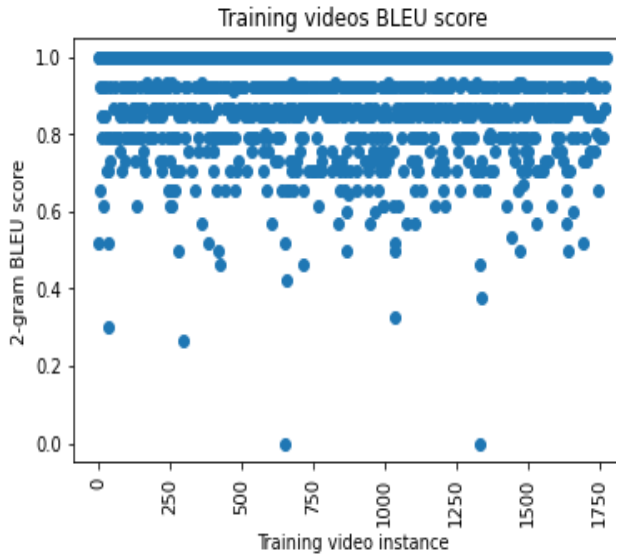

Fig. 8: Training BLEU scores.

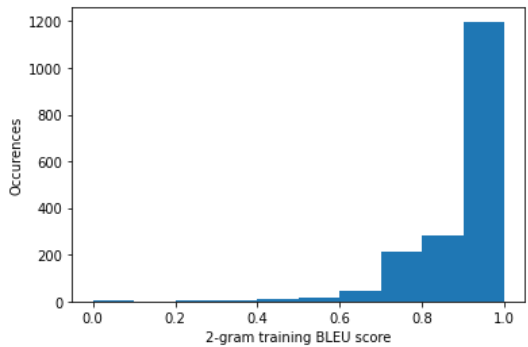

Fig. 9: Training BLEU histogram.

### B. Validation Set

For the validation set, the categorical accuracy of the model at the end of training was saturated at around 55%, so the expectation is that the BLEU scores between predicted and labeled captions in the validation set is worse than the training set. Figure 10 shows the BLEU scores across all videos in the validation set. Figure 11 shows the frequency distribution (histogram) of BLEU scores in the validation set. The average BLEU score for predicted captions on the training set is 45.8%.

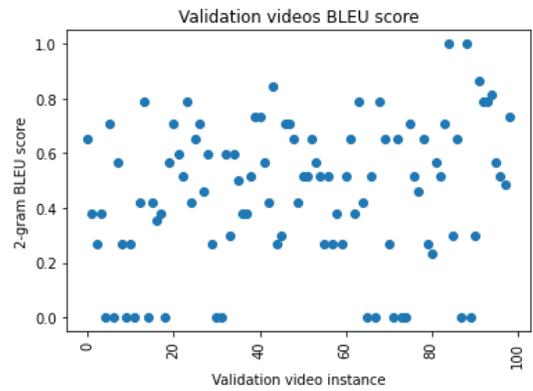

Fig. 10: Validation BLEU scores.

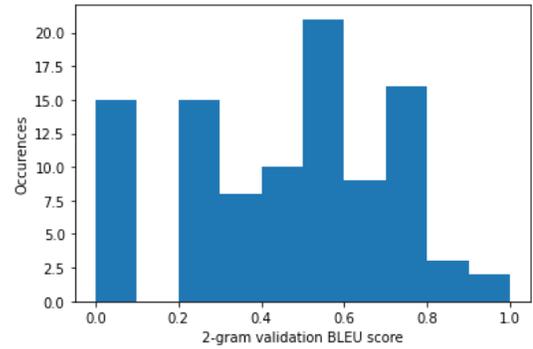

Fig. 11: Validation BLEU histogram.

### C. Test set

The test set is an identically random set to the validation set, so the expectation is the scores are the same in the test and validation set. Figure 12 shows the BLEU scores across all videos in the test set. Figure 13 shows the frequency distribution (histogram) of BLEU scores in the test set. The average BLEU score for predicted captions on the test set is 43.3%. The validation and test set are the same in that each dataset was not provided to the model for training. However, the validation set was used during training to assess loss and accuracy at each epoch, and was thus used for optimization. Therefore, holding out a separate test set used only during inference to assess model performance was necessary.

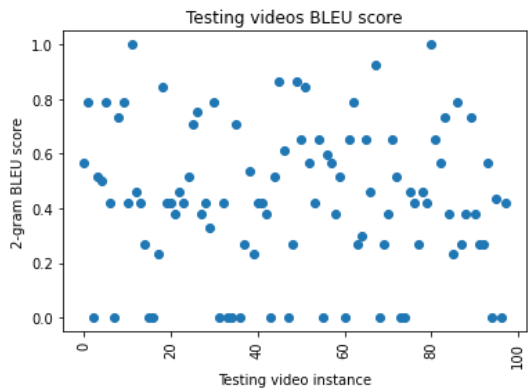

Fig. 12: Test BLEU scores.

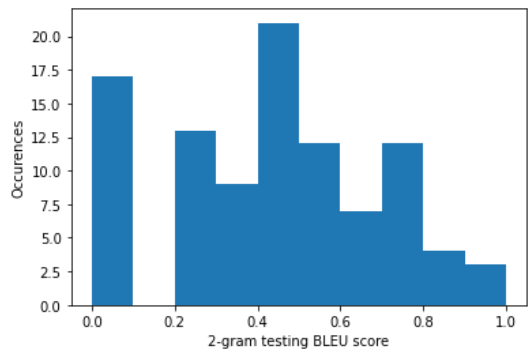

Fig. 13: Test BLEU histogram.

Figure 18 shows a video of a man cutting a water bottle with a sword. The labeled caption is "someone slices a bottle with a sword". However, the model predicted caption is "a man is cutting something with a tree". The model correctly captioned the first part of the video where the man cuts the bottle, and it also used the scene change in the last few frames of the video where a tree is in frame. Again it is clear the model is generalizing a caption across video frames.

## V. Discussions

The BLEU scores were computed using 2-gram similarity. Thus, the score was above zero even if only two-word sequences were similar between the predicted and groundtruth captions. For a stricter performance evaluation, a 5gram BLEU score could be used, as this is the midpoint of the sentence length of ten words. BLEU scores were above 90% for the training set, which demonstrates good model convergence for all videos and captions in the training set. Certainly, the model is complex enough for the given training data. The poorer performance of BLEU scores on the

### D. Caption prediction

To show the model's ability to generalize captions across video temporal frames compared to a single frame image caption, figures are presented here (as well as more in the appendix). In the following figures, the video feature matrix is displayed. Scene changes in the video appear in the feature matrix as changes throughout the vertical dimension (columns) of the matrix. One of the labeled descriptions (ground truth) for the video is displayed above the feature matrix. To show video scene changes across the video in this document, three frames are extracted at instances where the video scene is different. These frames were selected from rows in the feature matrix that show abrupt changes. The frames are labeled by number, which corresponds to the row in the feature matrix indicated by the red dashed horizontal line. Compelling videos are chosen from the dataset splits to report here.

Figure 14 shows a person wearing hockey equipment in the first 60 frames of the video, then the last 20 frames of the video switches to another scene with a baby sitting at a table. The labeled caption is "a baby is on football dress" while the predicted caption is "the baby dressed in equipment fell down over". The person wearing equipment in the beginning of the video indeed falls over, so this part of the caption is correctly predicted by the model. The model however uses the last few frames in which a baby is present to change the subject of the Figure 17 shows a video where a tortoise walks across sand, then changes scene to an otter swimming in water. The labeled caption is "a large tortoise is walking downhill" while the model predicted caption is "a small animal is running

The number of tokens in the vocabulary determines one dimension of the decoder model. Thus, increasing the words used for training will also increase the dimension of the one- validation and test data can be explained by model overtraining in the test set. The videos in the MSVD dataset have wildly varying subjects, including videos of sports, cooking, and other random actions. Thus, even though the videos were shuffled to generated the 3 splits in the dataset, one cannot expect the model to perfectly generalize to new video subjects. hot-encoded word vectors. With a latent dimension of 512 for each the model LSTMs, the number of trainable parameters of the model was already quite high at over 14 million. During training, the GPU memory of my training desktop used around 5.3gB out of the 6gB of the GTX 1660 Ti GPU. Thus, We were already dealing with memory issues loading the current model. We also had to keep the batch size at 50 video and sequence target pairs to ensure no memory issues. If a word is in the validation or test description sets, and it is not also in the training set, the BLEU score will be worse, as their will not be an n-gram match even if the sentence sentiment is the same. An analysis of words in the validation and test sets and their prevalence in the training vocabulary was not performed in this project, but certainly increasing the words from the training descriptions kept in the vocabulary will increase the likelihood that the training vocabulary is a superset of all words in the validation and test sets.

The sentence length was kept to ten words for the same memory performance as described in the previous . The prediction sentence length determines the second dimension of the text decoder input, so increasing this dimension will further increase the memory usage during model training. As the model is trained on videos with significant scene changes, there are often multiple subjects in a single video. Thus, there were videos for which the predicted caption was grammatically correct for the first ten words but incomplete. For example, Figure 15 shows a predicted caption "a girl is showing how to do a " which ends in the article "a". Thus, this sentence is grammatically correct and correctly describes the video subject but is incomplete due to the maximum predicted sentence length. As the video captioning model attempts to produce a general caption for all scenes in the video, videos with multiple subjects will require longer sentence to describe the multiple subjects. This performance could be improved by increasing the description length.

The most significant observation of the sequence-tosequence architecture for video captioning is that the predicted models are indeed able to generalize their captions across the video scene changes. Predicted captions were often more general and less specific in their description of the scenes. For videos where the scene dramatically changed to a different subject, the nouns present in both scenes were included in the output caption. For example, Figure 14 shows a video in the training set which switches dramatically to a different subject in later frames. The caption is "the baby dressed in equipment fell down over" correctly describes the action in the first scene, where a person wearing hockey equipment falls over. However, the last few frames show a baby sitting at a table and playing. The model correctly identifies the subject of the these frames, the baby, and inserts it into the action of the first scene of the video. Thus, it is clear that the model is able to utilize all of the temporal information in the frames of the video to produce a general caption, but this effect is at the sake of grammatical correctness of the predicted caption. Increasing the word length of the model (text decoder temporal dimension) would maintain the model generality but improve sentence grammar by allowing for compound sentences. A correct caption for this example could be "the person dressed in equipment fell over and a baby is sitting". These correct compound sentences certainly require a higher temporal dimension of the decoder LSTM. We believe this work can help most generative models like GPT, LLM, image generative model, and video generative model to interpret any visual data to get captions/texts.